\documentclass[letterpaper, 10 pt, conference]{ieeeconf}  

\IEEEoverridecommandlockouts                              

\usepackage{url}
\usepackage{graphicx}
\usepackage{amsmath}
\usepackage{amsfonts}
\usepackage{amssymb}

\usepackage{float}
\usepackage{cite}
\usepackage[caption=false,font=footnotesize]{subfig}

\title{\LARGE \bf
Extracting Traffic Primitives Directly from Naturalistically Logged Data for Self-Driving Applications
}

\author{Wenshuo Wang$^{1}$ and Ding Zhao$^{2}$
\thanks{$^{1}$Wenshuo Wang is with the Department of Mechanical Engineering, Beijing Insititue of Technology, Beijing, China 100081, and also with the Department of Mechanical Engineering, University of Michigan, Ann Arbor, MI, 48109 USA. He was with the Department of Mechanical Engineering, University of California at Berkeley, CA, 94720 USA.
        {\tt\small wwsbit@gmail.com}}%
\thanks{$^{2}$Ding Zhao is with the Department of Mechanical Engineering, University of Michigan, Ann Arbor, MI, 48109 USA.
        {\tt\small zhaoding@umich.edu}}%
}

\begin{document}

\maketitle
\thispagestyle{empty}
\pagestyle{empty}

\begin{abstract}

Developing an automated vehicle, that can handle complicated driving scenarios and appropriately interact with other road users, requires the ability to semantically learn and understand driving environment, oftentimes, based on analyzing massive amounts of naturalistic driving data. An important paradigm that allows automated vehicles to both learn from human drivers and gain insights is understanding the principal compositions of the entire traffic, termed as traffic primitives. However, the exploding data growth presents a great challenge in extracting primitives from high-dimensional time-series traffic data with various types of road users engaged. Therefore, automatically extracting primitives is becoming one of the cost-efficient ways to help autonomous vehicles understand and predict the complex traffic scenarios. In addition, the extracted primitives from raw data should 1) be appropriate for automated driving applications and also 2) be easily used to generate new traffic scenarios. However, existing literature does not provide a method to automatically learn these primitives from large-scale traffic data. The contribution of this paper has two manifolds. The first one is that we proposed a new framework to generate new traffic scenarios from a handful of limited traffic data. The second one is that we introduce a nonparametric Bayesian learning method -- a sticky hierarchical Dirichlet process hidden Markov model -- to automatically extract primitives from multidimensional traffic data without prior knowledge of the primitive settings. The developed method is then validated using one day of naturalistic driving data. Experiment results show that the nonparametric Bayesian learning method is able to extract primitives from traffic scenarios where both the binary and continuous events coexist.

\end{abstract}

\section{INTRODUCTION}
Autonomous vehicles play a vitally important role in intelligent traffic systems, road safety, and driving workload reduction\cite{wang2017human}. A lot of automated vehicle research has focused on how to learn end-to-end controllers\cite{yang17feature}, how to design and generate traffic scenarios for automated vehicle evaluation\cite{zhao2017evaluation}, how to understand the traffic scenes using naturalistic driving data based on deep learning and machine learning techniques, capable of offering supportive interventions to human drivers during a specific task. With this purpose, the automated vehicles need to fully understand how driving scenarios are changing and correctly predict what other road users surrounded will do. In such traffic scenarios where different road users are engaged, state changes of driving environment encompass the movements of both the automated vehicle and the other surrounding road users. When investigating traffic scenarios, researchers will manually and subjectively extract some specific scenarios they are interested from a handful of databases according to scenario definitions. These scenarios including car-following, lane-changing, overtaking behaviors, etc., however, are not able to cover the entire traffic case, and also not might be suitable to learn for algorithms. More specifically, these manually extracted scenarios are not flexible to be cascaded or combined to generate new traffic scenarios.

On the other hand, one of the greatest challenges also exists in manually extracting and reusing huge amounts of these scenarios because of dizzying databases and exploding data growth \cite{wang2017much}. Advanced sensing technologies such as cameras, radars, lidars, and GPS can provide rich information for modern vehicles\cite{erlien2016shared} (see \textsc{Appendix A}), enabling data-driven techniques to be one practicable way to deal with problematic issues in intelligent transportation systems\cite{zhang2011data}. Today, traffic in the real world is a vast and varied cyber-physical system, with thousands of kinds of drivers, vehicles and driving environment. The flood of data can overwhelm human insight and analysis because the size and complexity of traffic data sets are far larger and messier than a human being can manually cope with\cite{appenzeller2017scientists9}. Many autonomous companies and researchers put great efforts into collecting high quality and useful data and then annotating it, which is a greatly both time- and resource-consuming procedure, thus limiting the fast development of automated vehicles.  For example, DeepAI\cite{driveai2016} spends 800 human hours to label data for every one hour recorded data using deep learning techniques.

To obtain valuable and useful information from large databases, researchers developed powerful new technologies such as learning-based approaches. For example, Bender \textit{et al}. \cite{bender2015unsupervised} developed an unsupervised method, Bayesian multivariate linear model, to segment a time-series inertial data into finite amounts of linear portions for inferring driver behaviors, but only for a two-dimensional data sequence of one vehicle. Taniguchi \textit{et al}. \cite{taniguchi2016sequence} introduced a double articulation analyzer based on nonparametric Bayesian theory to predict driver behaviors with a six-dimensional data sequence, consisting of gas pedal position, brake pressure, steering angle, velocity, acceleration, and steering angle rate. They also developed an unsupervised approach to segment and predict driver's upcoming behavior by detecting and learning contextual changing points \cite{taniguchi2015unsupervised}. Hamada \textit{et al}. \cite{hamada2016modeling} applied a nonparametric Bayesian with linear dynamical systems to learn driver behavior primitives and predict drivers' upcoming behaviors. Wang \textit{et al}. \cite{wang2017drivingstyle} applied three different nonparametric Bayesian learning approaches to analyze human drivers' car-following styles. Learning-based methods have been widely used to model and predict driver behavior; however, according to our knowledge, no literature presents an insight into traffic scenario primitive extraction with surrounding vehicles engaged and regenerates to new traffic scenarios for automated vehicles.

\begin{figure}[t]
	\centering
	\vspace{0.1cm}
	\includegraphics[width = 0.46\textwidth]{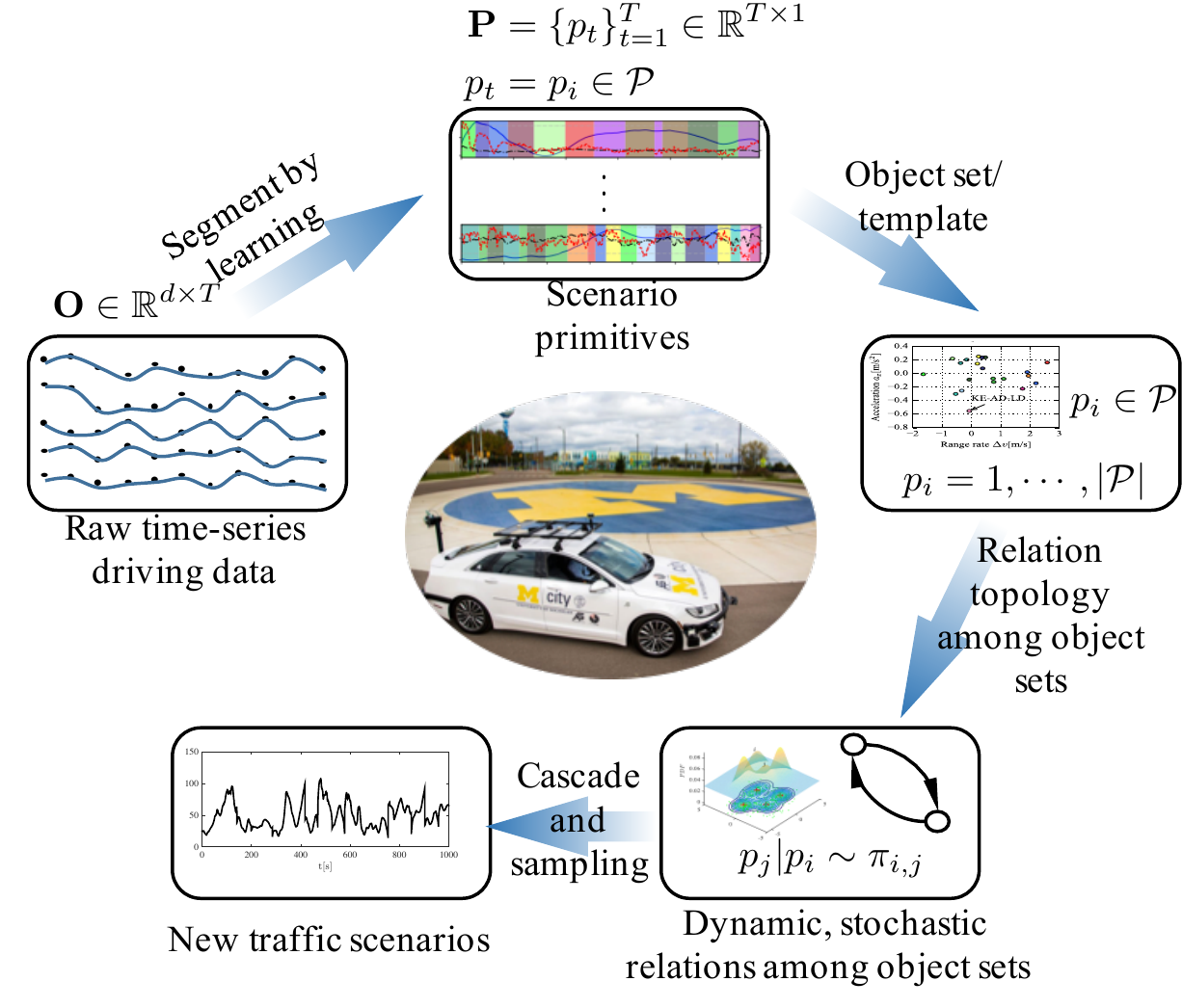}
	\caption{The proposed framework of generating new driving scenarios for automated driving applications. Three key attributes are involved in our proposed framework: compositionality, causality, and learning to learn, which is analogous to the human-level concept learning\cite{lake2015human}.}
	\label{fig:paperidea}
\end{figure}

Differing from previous research \cite{taniguchi2015unsupervised,taniguchi2016sequence,bender2015unsupervised,hamada2016modeling,yamazaki2016integrating} focusing on individual driver's behavior, we mainly concern how to generate an infinite number of new traffic scenarios from a handful of limited raw traffic data. Traffic scenarios could be generated by carefully and logically reshaping and cascading the basic compositions of traffic, termed as traffic primitive. Based on this, we proposed a primitive-based framework as shown in Fig. \ref{fig:paperidea}, which brings three key attributes -- \textit{compositionality}, \textit{causality}, and \textit{learning to learn}. The main tasks and challenges of achieving these are listed as follows:
\begin{enumerate}
	\item Developing algorithms that can automatically extract these primitives without prior knowledge about the type and number of them.
	\item Finding and clustering the analogous primitives, thus generating the object set or template.
	\item Describing and modeling the topological relations between template sets, and then obtaining a causal structure for dynamic, stochastic relations.
	\item Proposing a method to automatically generate infinite amounts of new scenarios being statistically equivalent to what the vehicle would have encountered in real life.
\end{enumerate}
We note that, the first two challenges are to parse a long-term, multiscale time-series driving data into primitives and cluster them. To achieve this, we introduce a nonparametric Bayesian learning method to extract the traffic scenario primitives from a traffic data sequence. In the real traffic scenarios, however, two kinds of events primarily exist, one is described using binary states and the other one is described using continuous states. The binary state represents a new road user's appearance/disappearance in current driving scenarios and the continuous state represents the state changes of current driving scenarios with a fixed number of surrounding road users involved. This paper presents the contributions as follows.
\begin{itemize}
	\item Presenting a primitive-based framework capable of generating new traffic scenarios for self-driving applications.
	\item Implementing a nonparametric approach to extract traffic primitives and demonstrating its utility on multiscale traffic data.
\end{itemize}

The remains of this paper are organized as follows. Section II introduces the developed nonparametric Bayesian learning method. Section III presents the experiment procedure and data collection. Section IV shows the experiment results and analysis. Finally, the conclusions are given in Section V.

\section{METHODS}

\begin{figure}[t]
	\centering
	\subfloat[HMM]{\includegraphics[width = 0.35\textwidth]{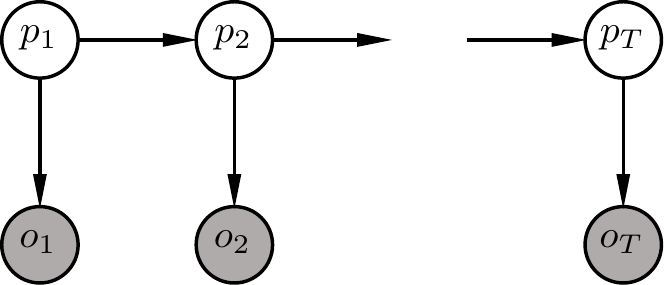}}\\
	\subfloat[sticky HDP-HMM]{\includegraphics[width = 0.23\textwidth]{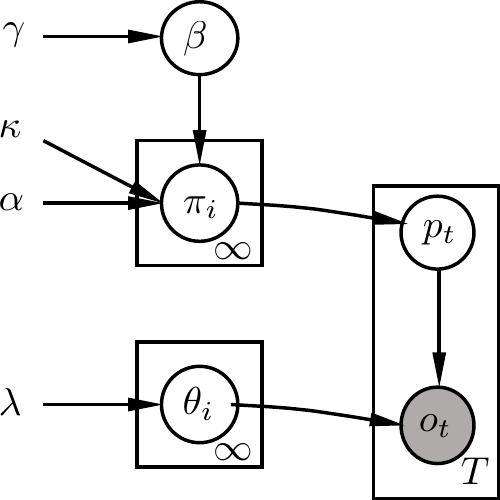}}
	\caption{Graphic illustration of HMM and sticky HDP-HMM.}
	\label{fig:HMM-HDP-HMM}
\end{figure}

This section will detail a sticky HDP-HMM method, which is being powerful to model driver behaviors in the case where the number of primitive driving patterns is not exactly known. In what follows, the theoretical basis of hidden Markov model (HMM) and hierarchical Dirichlet processes (HDP).
\subsection{HMM}
When facing the uncertainty of driver behaviors in naturalistic settings, we can treat the entire driving process as a logic combination of primitives $p\in \mathcal{P}$, and the dynamic process among primitives in driver behaviors as a probabilistic inferential process\cite{donoso2014foundations}. Here, driver behaviors were modeled as a dynamic process of primitives with the structure of HMM. The core of HMM consists of two layers: a layer of hidden \textit{state}, representing the driving primitives and a layer of \textit{observation}, as shown in Fig. \ref{fig:HMM-HDP-HMM}(a).

Given an observed time-series data sequence $ \mathbf{O} = \{o_{t}\}_{t=1}^{T} $ with $ \mathbf{O}\in \mathbb{R}^{d\times T} $ and a set of primitive $ \mathcal{P} $, each driving primitive $ p_t $ at time $ t $ will be subject to one entry of $ \mathcal{P} $, i.e., $ p_{t} = p_{i}\in \mathcal{P} $, where $ p_i $ is the $ i $-th element in $ \mathcal{P} $. The transition probability from primitive $ p_{i} $ to $ p_{j} $ is denoted as $ \pi_{i,j} $, and $ \pi_{i} = [\pi_{i,1}, \pi_{i,2}, \pi_{i,3}, \cdots] $. The observation $ o_{t} $ at time $ t $ given primitive $ p_{t} $ is generated by $ o_{t} = F(o_{t}|p_{t}) $, called emission function. Therefore, the HMM can be described as
\begin{subequations}
	\begin{align}
	p_{t}|p_{t-1} & \sim \pi_{p_{t-1}} \\
	o_{t}| p_{t} & \sim F(\theta_{p_{t}})
	\end{align}
\end{subequations}
where $ F(\cdot) $ is the emission function and $ \theta_{p_{t}} $ is the emission parameter. Driver behavior, however, are changing and open-ended,  so that the parameter space regarding hidden states in the model (i.e., the number of primitives, the size of transition matrix) becomes potentially infinite\cite{donoso2014foundations}.  More specifically, the dimension of the set space of driving primitive, $ |\mathcal{P}| $, is unknown. In such situations, we have to define a prior probability distribution on an infinite-dimensional space.  A distribution on an infinite-dimensional space is a stochastic process with a specific path. Usually, the Dirichlet processes (DP) rapidly yield intractable computations. In what follows, we will introduce a hierarchal DP (HDP).

\subsection{HDP}
We assume that the type of driving primitive in (1) is priorly unknown and these primitives in HMM are subject to a specific distribution defined over a measure space. The Dirichlet process (DP) is a measure on measures, denoted by DP($ \gamma, H $), and provides a distribution over discrete probability measures with an infinite collection of atoms
\begin{subequations}
	\begin{align}
	G_{0} & = \sum_{i=1}^{\infty}\beta_{i}\delta_{\theta_{i}}, \ \ \ \ \theta \sim H \\
	\beta_{i} & = \nu_{i} \prod_{\ell = 1}^{i-1}(1-\nu_{\ell}), \ \ \ \ \nu_{i} \sim \mathrm{Beta}(1, \gamma)
	\end{align}
\end{subequations}
on a parameter space $ \Theta $ that is endowed with a base measure $ H $. Here, the weights $ \beta_{i} $ sampled by a stick-breaking construction and we denote $ \beta \sim \mathrm{GEM}(\gamma) $, with $ \beta = [\beta_{1}, \beta_{2}, \beta_{3}, \cdots] $ and $ \sum_{i}^{\infty} \beta_{i} = 1 $.

According to the above discussion, an HDP can be used to define a prior on the set of HMM transition probability measures $ G_{j,i} $
\begin{equation}
G_{j,i} = \sum_{i = 1}^{I}\pi_{j,i}\delta_{\theta_{i}}
\end{equation}
where $ \delta_{\theta} $ is a mass concentrated at $ \theta $. Assuming that each discrete measure $ G_{j} $ is a variation on a global discrete measure $ G_{0} $, thus the Bayesian hierarchical specification takes $ G_{j} \sim \mathrm{DP}(\alpha, G_{0}) $, where $ G_{0} $ is draw from $ \mathrm{DP}(\gamma, H) $:
\begin{subequations}\label{equation:DP}
	\begin{align}
	G_{0} & = \sum_{i=1}^{\infty}\beta_{i}\delta_{\theta_{i}}, \ \ \ \ \beta|\gamma \sim \mathrm{GEM}(\gamma) \\
	G_{j} & = \sum_{i=1}^{\infty}\pi_{j,i}\delta_{\theta_{i}}, \ \ \ \ \pi_{j}|\alpha, \beta \sim \mathrm{DP}(\alpha,\beta)\\
	\theta_{i}|H & \sim H
	\end{align}
\end{subequations}
where $\mathrm{GEM}(\gamma)$ is a distribution, named after Griffiths, Engen and McCloskey.
\subsection{Sticky HDP-HMM}
For the sticky HDP-HMM($ \gamma,\alpha,H $), by adding an extra parameter $ \kappa > 0 $ that biases the process toward self-transition in (\ref{equation:DP}b), increasing the expected probability of self-transition by an amount proportional to $ \kappa $, we can obtain
\begin{subequations}\label{equation:sticky_HDP_HMM}
	\begin{align}
	\beta|\gamma&\sim \mathrm{GEM}(\gamma) \\
	\pi_{i}|\alpha,\beta,\kappa&\sim \mathrm{DP}(\alpha + \kappa, \frac{\alpha\beta + \kappa \delta_{i}}{\alpha + \kappa}), \ i = 1, 2, \cdots \\
	p_{t}|p_{t-1}&\sim \pi_{p_{t-1}}, \ t = 1,2, \cdots, T  \\
	o_{t}|p_{t},\theta_{p_{t}}&\sim F(\theta_{p_{t}}), \ t = 1,2, \cdots, T \\
	\theta_{i}|H&\sim  H, \ i = 1,2, \cdots.
	\end{align}
\end{subequations}
where $ T $ is the data length. Note that when $ \kappa = 0 $ in (\ref{equation:sticky_HDP_HMM}b), the original HDP-HMM is obtained.

\subsection{Emission Model}
The observation model is determined by the type of function $ F(\theta_{i}) $, which can be Gaussian emissions\cite{mahboubi2017learning} or switch linear dynamic models (SLDSs)\cite{fox2011bayesian} (e.g., vector autoregressive). One main challenge with non-parametric approaches is that one must derive all the necessary expressions to properly perform inference\cite{ryden2008versus}. Here, to make our algorithm tractable, we assume that observations are drawn from a Gaussian distribution like in \cite{mahboubi2017learning}. The $ \theta_{i} $ is set as $ \theta_{i} = [\mu_{p_i}, \Sigma_{p_i}] $. Therefore, if the priors for observations and transition distributions are learned correctly, the full-conditional posteriors can be computed using Gibbs sampling method. Johnson and Willsky\cite{johnson2013bayesian} present further details of the inference method using Gibbs sampling methods.

\section{EXPERIMENT AND DATA COLLECTION}

\begin{figure}[t]
	\centering
	\vspace{0.1cm}
	\includegraphics[width = 0.45\textwidth]{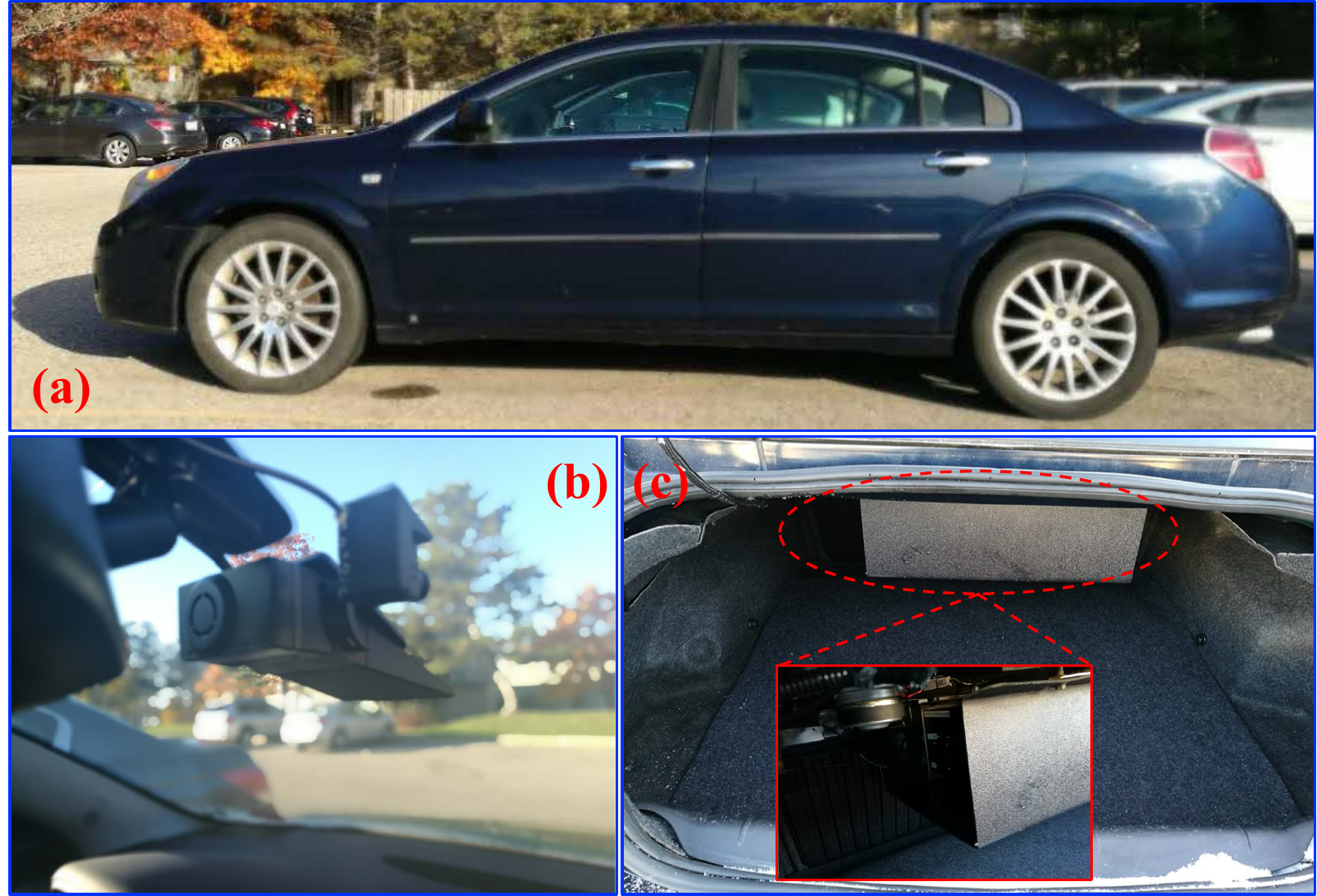}
	\caption{An example of equipped vehicles in our experiment. (a) Equipped vehicle; (b) Mobileye; (c) data collection systems}
\end{figure}

\subsection{Experiment Procedure}
Driving data used in this paper are extracted from the Safety Pilot Model Deployment (SPMD) database logged in Ann Arbor, Michigan. We use the equipped vehicles to run experiments and collected on-road data. The experiment vehicles are equipped with data acquisition systems and Mobieye. The road information (e.g., lane width, lane curvature) and the surrounding vehicle's information (e.g., relative distance, relative speed) are recorded by Mobileye. The subject vehicle information such as speed, steering angle, acceleration/brake pedal position is extracted from CAN-bus signal\cite{zhao2017trafficnet}. All of the data are recorded at 10 Hz.

Drivers had an opportunity to become accustomed to the equipped vehicles. They performed casual daily trips for several months without any restrictions on or requirements for their trips, the duration of the trips, or their driving style. The data processing and recording equipment were hidden from the drivers, thus avoiding the influence of recorded data on driver behavior.

\begin{figure}[t]
	\centering
	\includegraphics[width = 0.45\textwidth]{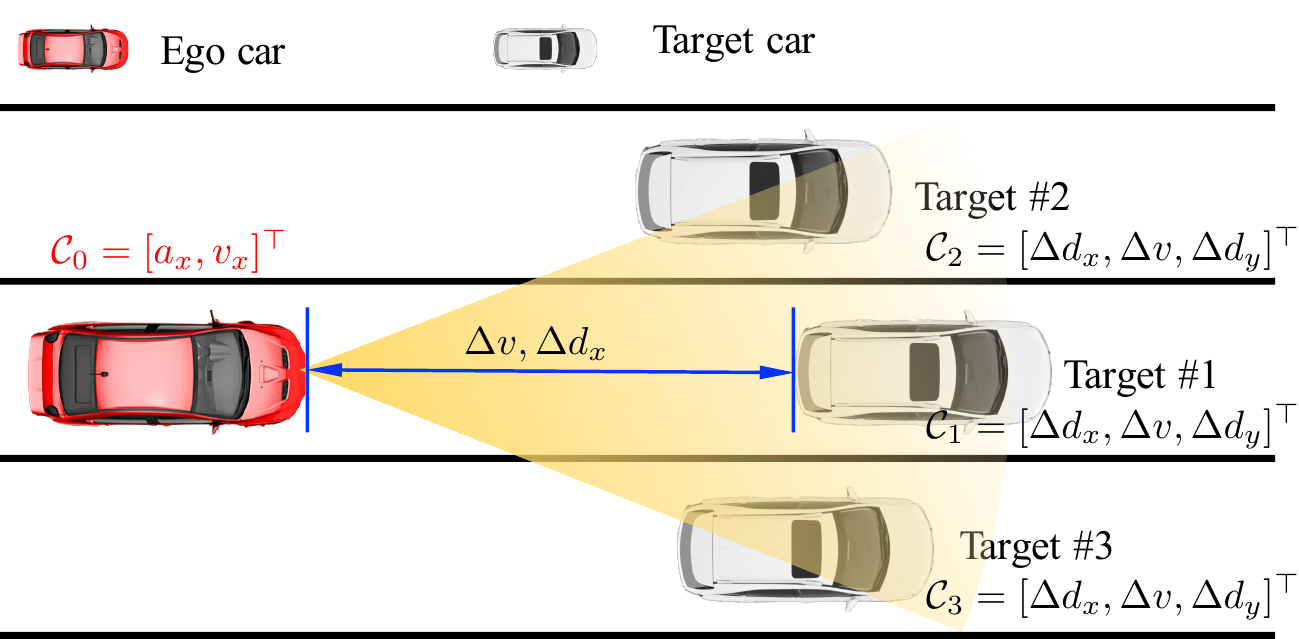}
	\caption{Driving scenarios consisting of surrounding vehicles for data collection.}
	\label{fig:driving_scene}
\end{figure}

\subsection{Data Collection}
In this work, we consider the driving scenarios where the ego vehicle can sense vehicles in front of it using Mobileye, as shown in Fig. \ref{fig:driving_scene}. In order to describe the data sequence easily, we define a \textit{channel}, $ \mathcal{C}_{k}^{T} $, to record the data sequence over time $ t = 1,2, \cdots, T $ from a single target car $ k $, where $ k = 1,\cdots,K$. Here, we set $ K=5 $ since the maximum amount of target cars that Mobileye can detect in front of the ego car is $ 5 $. In each channel, the data at time $ t $ recorded from each target car consists of three variables:
\begin{itemize}
	\item $ \Delta d_x^{t}$, relative distance (range) ;
	\item $ \Delta v^{t} $, relative range rate (i.e., relative speed);
	\item $ \Delta d_y^{t} $, lateral displacement of target cars with respect to lane boundary.
\end{itemize}
For each channel, we have $ \mathcal{C}_{k}^{T} = [\Delta d_x^{1:T}, \Delta v^{1:T}, \Delta d_y^{1:T}]^{\top} \in \mathbb{R}^{3\times T} $. For all channels, we initially set all variables to zeros. When the Mobileye detect the appearance of a target car in front of the ego car, the corresponding channel was activated and then recorded data. If the target car disappears in the detection region of Mobileye, the data in this channel was set to zero again.

Note that the collected data sequence contains two types of information or events -- binary and continuous:
\begin{enumerate}
	\item For the binary event, it records the appearance and disappearance of a target car in front of the ego car. The value in $ \mathcal{C}_{k} $ will usually be in form of step signal, which means that the cut-in or cut-out behavior of target cars in front of the ego car can be detected.
	\item In the continuous event, it records the target cars' states, the ego car's states (e.g., longitudinal speed, $ v_{x} $ and acceleration, $ a_x $) and their relative dynamic states (i.e., $ \Delta d_x, \Delta v, \Delta d_y $) when no target car cut-in or cut-out.
\end{enumerate}
In addition, we also use an additional channel, $ \mathcal{C}_{0} $, to record the ego vehicle's states, with $ \mathbf{O} = \mathcal{C}_{0}^{T} = [a_{x}^{1:T}, v_{x}^{1:T}]^{\top} \in \mathbb{R}^{2\times T}$. It is obvious that the channel $ \mathcal{C}_{0}^{T} $ will always record the continuous events. Totally, a data sequence $ \mathcal{C}^{T} $ with dimension $ 3K+2 $ is recorded from $ K $ target cars and one ego car, i.e., $ \mathcal{C}^{T} = [\mathcal{C}_{0}^{T}, \{\mathcal{C}_{k}^{T}\}_{k=1}^{K}] \in \mathbb{R}^{(3K+2)\times T} $. The developed model in this paper should be able to extract primitives from not only the binary events but also the continuous events.

\begin{table}[t]
	\centering
	\caption{\textsc{Parameter Values for Models}}
	\begin{tabular}{ccc}
		\hline
		\hline
		parameter & description  & value\\
		\hline
		$ (a_{\alpha}, b_{\alpha}) $ & $ \alpha $ gamma prior  & (1,1) \\
		$(a_{\gamma}, b_{\gamma})$ & $ \gamma $ gamma prior & (1,1) \\
		$ (a_{\kappa}, b_{\kappa}) $ & $ \kappa $ gamma prior& (100,1)\\
		$ n_{0} $ & IW prior degree of freedom & $ d+2 $ \\
		$ S_{0} $ & IW prior scale & 0.75$\cdot \bar{\Sigma} $\\
		\hline
		\hline
	\end{tabular}
	\label{Table:parameters}
\end{table}

\begin{figure}[t]
	\centering
	\subfloat[]{\includegraphics[width = 0.20\textwidth]{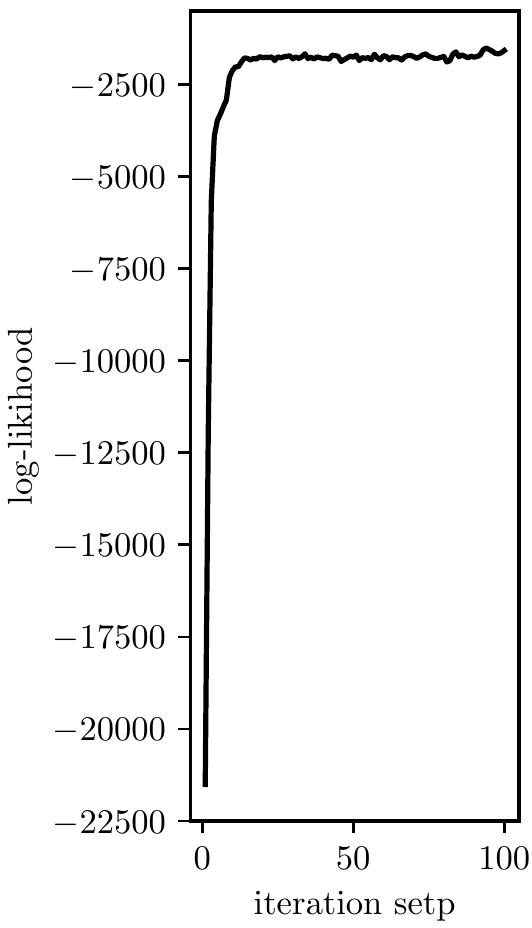}}
	\subfloat[]{\includegraphics[width = 0.19\textwidth]{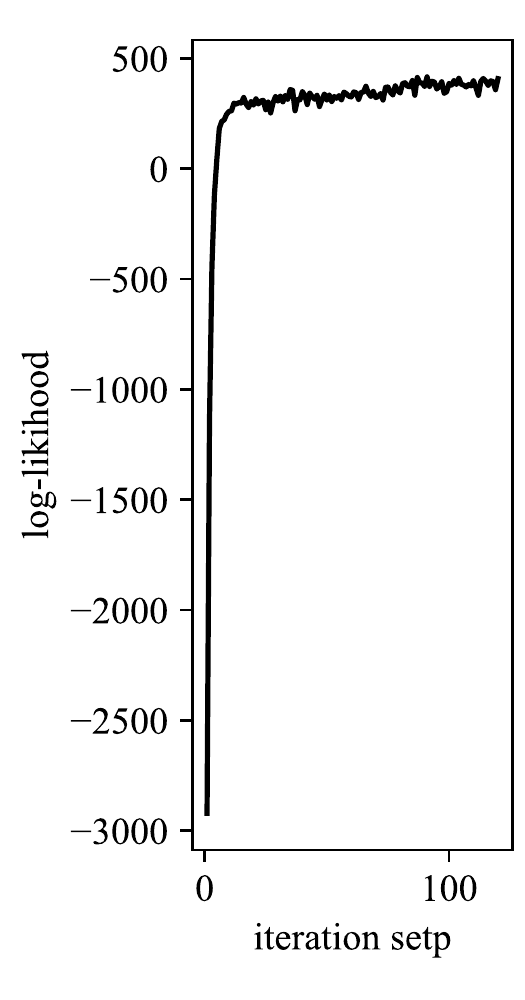}}
	\caption{The log-likelihood values with respect to the iteration steps for dealing with (a) binary events and (b) continuous events.}
	\label{fig:loglikelihood}
\end{figure}

\begin{figure*}[t]
	\centering
	\vspace{0.1cm}
	\includegraphics[width = 0.92\textwidth]{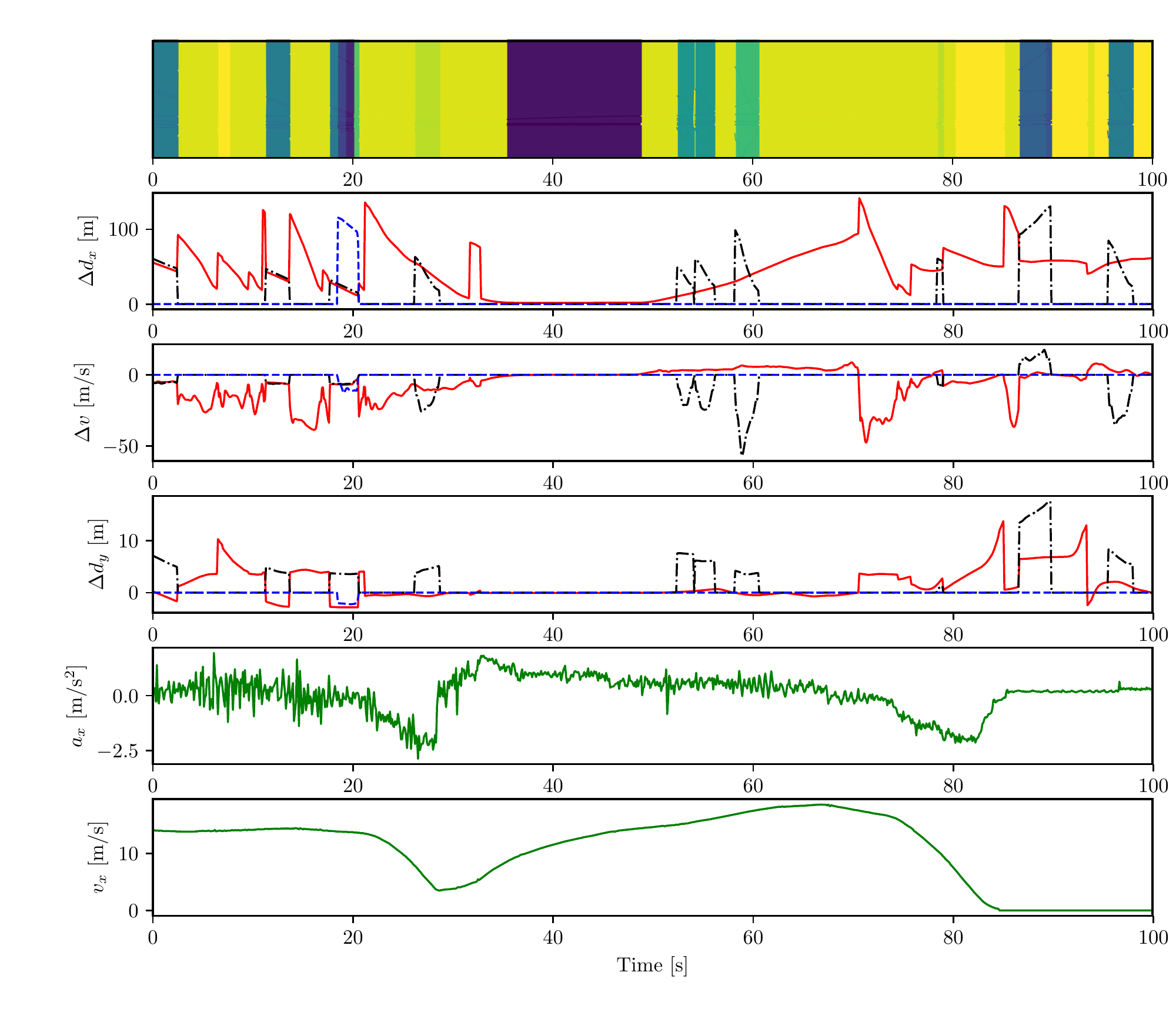}
	\caption{Example of experiment results for daily driving data with five channels, $ T = 100 $ s, and 12 kinds of primitives. Green line: channel \#0,  $ \mathcal{C}_{0}^{T} $; red line: channel \#1, $ \mathcal{C}_{1}^{T} $; black dot line: channel \#2; $ \mathcal{C}_{2}^{T} $; and blue line: channel \#3, $ \mathcal{C}_{3}^{T} $; signals in channel \#4 and channel \#5 are always being zero since no target car appeared in this channel for this example. Channels $ \mathcal{C}_{k}^{T} $, $ k = 1,2,3 $ consist of three variables, i.e., $ \Delta d_x $, $ \Delta v $, $ \Delta d_y $. The step signal means that a target car appeared or disappeared in Mobileye's sensing region.}
	\label{fig:binary}
\end{figure*}

\subsection{Training Procedure}
We develop and test the developed models based on Johnson and Willsky's \cite{johnson2013bayesian} as well as Fox's \cite{wulsin2014modeling} previous work using Python. The hyperparameters are determined using following rules:

\begin{enumerate}
	\item We place a Gamma($ a,b $) conjugate prior on the hyperparameters $ \gamma, \alpha, \kappa $ to make posterior estimation tractable as shown in Table \ref{Table:parameters}, where $ d $ is the dimension of input data, and set $ d = 3K+2 $ in this work.
	\item The Inverse-Wishart (IW) prior is conjugate to the Gaussian distributions, thus the hyperparameters for $ \theta_{i} $ are taken to be from an IW with a hyper-parameter $ \gamma $, i.e.,
	
	\begin{equation*}
	\Sigma_{i}|n_0, S_0 \sim \mathrm{IW}(n_0, S_0)
	\end{equation*}
	where $ n_0 $ is IW prior degree of freedom and $ S_0 = 0.75 \bar{\Sigma} $ is the IW prior scale with the covariance ($ \bar{\Sigma} $) of the observed data.
\end{enumerate}
In this work, the observation variables are generated from a Gaussian model and we set $ \mu_{p_i} = 0 $ according to \cite{hamada2016modeling}. For the case of binary events, we test and evaluate the method performance in the daily traffic scenarios where the appearance/disappearance of target cars will be involved. For the case of continuous events, we evaluate the method performance in the primitives extracted from daily traffic scenarios. Fig. \ref{fig:loglikelihood} gives the log-likelihood of learning results with respect to the iteration steps for dealing with binary and continuous events.

\begin{figure*}[t]
	\centering
	\vspace{0.1cm}
	\includegraphics[width = 0.85\textwidth]{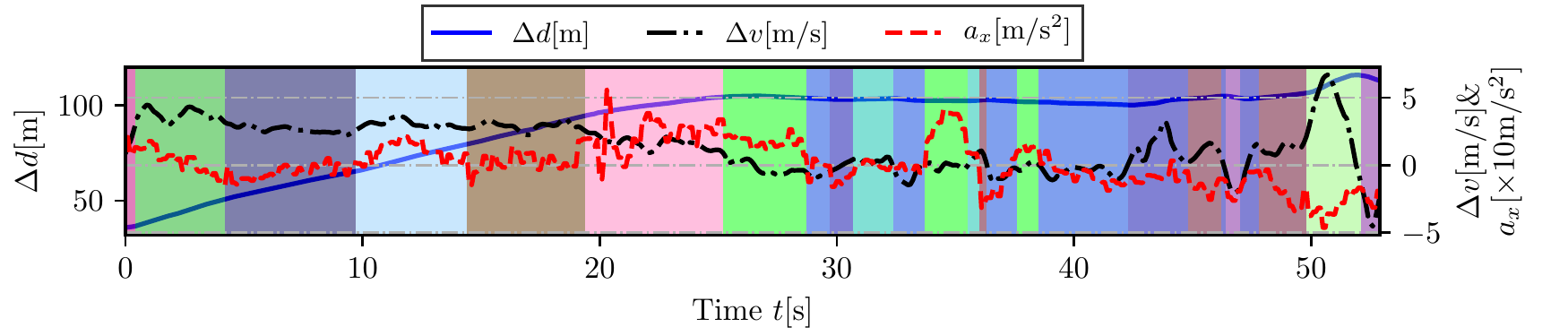}
	\caption{Example of experiment results for primitives with continuous events extracted from the daily driving data using the developed method, and resulting in 14 primitives.}
	\label{fig:primitive_con}
\end{figure*}

\section{RESULTS AND ANALYSIS}
For the developed method, we will evaluate its utility based on the ability to extract primitive from binary and continuous events in time-series sequences.

\subsection{Binary Event}
Regarding the binary event, we evaluate the utility by checking whether the proposed method can detect the appearance and disappearance of heading target cars. The ground truth can be obtained from the changes of target cars' label. Fig. \ref{fig:binary} shows an example of the learning primitive extraction results. The fact that the signal step points (i.e., appearance/disappearance of target cars) are detected indicates that the sticky HDP-HMM extracts traffic scenario primitives from a high dimensional (in this paper, the dimension of data sequence is $ d = 17 $) with different types of variables, though few points are not extracted such as at time $ t  = 31.6 $ s and $ t = 70.6 $ s. Also, the developed approach can cluster the primitives possessing the same attributes, that is, primitives with the same color have been assigned to the same label. For the 100s-data sequence, we finally obtained 12 primitives.

\subsection{Continuous Event}
There is no ground truth for the primitives extracted from continuous variables, that is, we would not subjectively and manually define the length of each primitive. Take speed for example, we would not empirically set a subjective threshold to segment speed profiles due to the variances in the speed profile among drivers\cite{wang2017driving}. Here, we learn the primitives using the sticky HDP-HMM, which can automatically find the primitive edges and then assign labels to each primitive. Fig. \ref{fig:primitive_con} presents an example of primitive extraction results for multidimensional continuous variables without step signal. We note that the sticky HDP-HMM can automatically learn primitives and assign the primitives with similar attributes to the same cluster, labeled with the same color.

\subsection{Statistical Results for Primitives}
\begin{figure}[t]
	\centering
	\includegraphics[width = 0.48\textwidth]{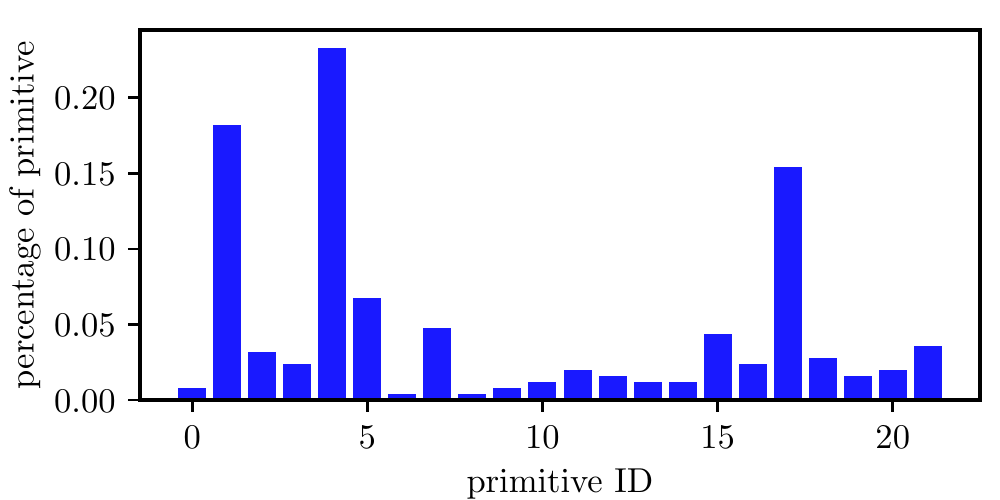}
	\caption{The statistical results of primitives of binary events with one day driving data for one driver.}
	\label{fig:bin_stat}
\end{figure}

In order to show the utility of the sticky HDP-HMM, we also give the statistical results for primitives. Primitives with the same color have been labeled to the same ID, called primitive ID. Fig. \ref{fig:bin_stat} presents the statistical results of learned primitives for one driver, with 353 primitives in total and 22 kinds of primitives. The horizontal axis is the primitive ID and the vertical axis is the percentage of each primitive. We note that the sticky HDP-HMM can automatically extract traffic scenario primitives from multiscale traffic database and then assign primitives endowing with the same attributes to one cluster. Each primitive ID indicates a primitive set which consists of a varying number of primitives. Table \ref{Table:extraction} lists the experiment results of  one day driving data for ten drivers with a high dimension at $ d = 14 $ and $ d = 17 $. The experiment results demonstrate that the introduced nonparametric Bayesian learning method can be applied to high-dimension and large time-scale data sequences.

\begin{table}[t]
	\centering
	\caption{Extraction Results of Ten Cars for Binary Events.}
	\begin{tabular}{ccccc}
		\hline
		\hline
		Vehicle ID & primitive sets & total primitives & $ T $ [s] & $ d $\\
		\hline
		10106 & 18 & 412 & 1046 & 14 \\
		10116 & 22 & 827 & 1990 & 17 \\
		10120 & 24 & 826 & 1302 & 17 \\
		10121 & 23 & 1090 & 2858 & 14 \\
		10122 & 15 & 425 & 2194 & 14\\
		10131 & 21 & 469 & 1149 & 17 \\
		10135 & 23 & 1178 & 2161 & 17 \\
		10137 & 26 & 702 & 3131 & 17 \\
		10145 & 31 & 793 & 1912 & 17 \\
		10154 & 23 & 931 & 1567 & 17 \\
		\hline
		\hline
		\multicolumn{5}{l}{Here, $ d = 14$ indicates that only 4 target cars were detected during}\\
		\multicolumn{5}{l}{ experiment.}
	\end{tabular}
	\label{Table:extraction}
\end{table}

\section{CONCLUSIONS AND FUTURE WORK}
In this paper, we proposed a new framework to generate an infinite number of new traffic scenarios with a handful of limited raw traffic data, consisting of four steps: primitive extraction, learning primitive sets, topology modeling between primitive sets, and generate new traffic scenarios using primitives. To achieve this, we introduced a nonparametric Bayesian learning method to deal with the challenges in the first two steps, i.e., extracting primitives from multiscale traffic scenarios, where the binary and continuous events are both involved, and obtain the object sets. The experiment results show that the introduced method can automatically obtain the primitives for binary events that encompass distinct primitive edges and also segment continuous events being without recognizable primitive edges. Also, this approach can also automatically cluster the primitives. The introduced nonparametric Bayesian learning approach enables one to extract primitives from a huge amount of multiscale time-series traffic data in a low cost of time and resources. 

This paper shows a sticky HDP-HMM approach to deal with the challenges in the first two steps for generating an infinite number of new traffic scenarios. Also, the primitive extraction can be used to analyze, model, and predict driver behaviors\cite{wang2017drivingstyle,hamada2016modeling}. The propose framework in Fig. \ref{fig:paperidea} can also be applied to robotics or human behavior analysis. The developed approach shows a satisfied ability to automatically segment different kinds of time-series data including both binary and continuous cases, however, the extracted primitives still could not be fully interpretable. Making extracted primitives interpretable could provide us insights into the complex traffic scenarios. We will develop a human-inspired approach which can automatically extract interpretable primitives. In addition, the distinct levels among the extracted primitives are still not quantitatively clear. Many existed approaches can be directly used to comprehensively evaluate the distinct levels of different types of driving primitive clusters, for example, by measuring the similarity levels between cluster distributions using Kullback-Leibler divergence \cite{wang2017drivingstyle} or by comparing with results of using clustering methods\cite{mahboubi2017learning,grigorediscovering}.

\addtolength{\textheight}{-12cm}   



\section*{APPENDIX}
Ten released databases are listed as follows:
\begin{itemize}
	\item KITTI Vision Benchmark Suite\footnote[1]{\url{http://www.cvlibs.net/datasets/kitti/index.php}}
	\item Vision for Intelligent Vehicles and Applications\footnote[2]{\url{http://cvrr.ucsd.edu/vivachallenge/}}
	\item Oxford RobotCar Dataset\footnote[3]{\url{http://robotcar-dataset.robots.ox.ac.uk/}}
	\item The University of Michigan North Campus Long-Term Vision and LIDAR Datasets\footnote[4]{\url{http://robots.engin.umich.edu/nclt/}}
	\item DIPLECS Autonomous Driving Datasets\footnote[5]{\url{http://www.diplecs.eu/index.html}}
	\item Velodyne SLAM  Dataset\footnote[6]{\url{http://www.mrt.kit.edu/z/publ/download/velodyneslam/dataset.html}}
	\item SYNTHIA Dataset\footnote[7]{\url{http://synthia-dataset.net/}}
	\item Daimler Urban Segmetation Dataset \footnote[8]{\url{http://www.6d-vision.com/scene-labeling}}
	\item MIT Age Lab\footnote[9]{\url{http://lexfridman.com/carsync/}}
	\item MOLP dataset\footnote[10]{\url{http://hcr.mines.edu/code/MOLP.html}}
\end{itemize}

\section*{ACKNOWLEDGMENT}
Toyota Research Institute ("TRI") provided funds to assist the authors with their research but this article solely reflects the opinions and conclusions of its authors and not TRI or any other Toyota entity.
%



\bibliographystyle{IEEEtran}
\bibliography{Extraction}

\end{document}